\documentclass[letterpaper]{article} % DO NOT CHANGE THIS
\usepackage{aaai25}  % DO NOT CHANGE THIS
\usepackage{times}  % DO NOT CHANGE THIS
\usepackage{helvet}  % DO NOT CHANGE THIS
\usepackage{courier}  % DO NOT CHANGE THIS
\usepackage[hyphens]{url}  % DO NOT CHANGE THIS
\usepackage{graphicx} % DO NOT CHANGE THIS
\usepackage{natbib}  % DO NOT CHANGE THIS AND DO NOT ADD ANY OPTIONS TO IT
\usepackage{caption} % DO NOT CHANGE THIS AND DO NOT ADD ANY OPTIONS TO IT
\usepackage{amsmath}
\usepackage{amssymb}
\usepackage{algorithm}
\usepackage{algorithmic}
\usepackage{bm}
\usepackage{bbding}
\usepackage{booktabs}
\usepackage{enumitem}
\usepackage{multirow}
\usepackage{makecell}
\usepackage{subfig}

\usepackage{newfloat}
\usepackage{listings}
\DeclareCaptionStyle{ruled}{labelfont=normalfont,labelsep=colon,strut=off} % DO NOT CHANGE THIS
\floatstyle{ruled}
\newfloat{listing}{tb}{lst}{}
\floatname{listing}{Listing}

\title{Implicit Location-Caption Alignment via Complementary Masking \\ for Weakly-Supervised Dense Video Captioning}

\author{
Shiping Ge\textsuperscript{\rm 1}\thanks{Work done during an internship at Tencent WeChat.}, 
Qiang Chen\textsuperscript{\rm 2}, 
Zhiwei Jiang\textsuperscript{\rm 1}\thanks{Corresponding author.}, 
Yafeng Yin\textsuperscript{\rm 1}, 
Liu Qin\textsuperscript{\rm 2}, 
Ziyao Chen\textsuperscript{\rm 2}, 
Qing Gu\textsuperscript{\rm 1} 
}

\affiliations{
    \textsuperscript{\rm 1}State Key Laboratory for Novel Software Technology, Nanjing University, Nanjing, China \\
    \textsuperscript{\rm 2}Tencent WeChat, Guangzhou, China \\
    shipingge@smail.nju.edu.cn, ethanqchen@tencent.com, \{jzw, yafeng\}@nju.edu.cn, \\ \{stenliu, yateschen\}@tencent.com, guq@nju.edu.cn
}

\begin{document}

\maketitle

\begin{abstract}
Weakly-Supervised Dense Video Captioning (WSDVC) aims to localize and describe all events of interest in a video without requiring annotations of event boundaries. 
This setting poses a great challenge in accurately locating the temporal location of event, as the relevant supervision is unavailable.  
Existing methods rely on explicit alignment constraints between event locations and captions, which involve complex event proposal procedures during both training and inference.
To tackle this problem, we propose a novel implicit location-caption alignment paradigm by complementary masking, which simplifies the complex event proposal and localization process while maintaining effectiveness. 
Specifically, our model comprises two components: a dual-mode video captioning module and a mask generation module. 
The dual-mode video captioning module captures global event information and generates descriptive captions,
while the mask generation module generates differentiable positive and negative masks for localizing the events.
These masks enable the implicit alignment of event locations and captions by ensuring that captions generated from positively and negatively masked videos are complementary, thereby forming a complete video description.
In this way, even under weak supervision, the event location and event caption can be aligned implicitly.
Extensive experiments on the public datasets demonstrate that our method outperforms existing weakly-supervised methods and achieves competitive results compared to fully-supervised methods. 

\end{abstract}

\begin{links}
    \link{Code}{https://github.com/ShipingGe/ILCACM}
\end{links}

\begin{figure}[t]
    \centering
    \includegraphics[width=\linewidth]{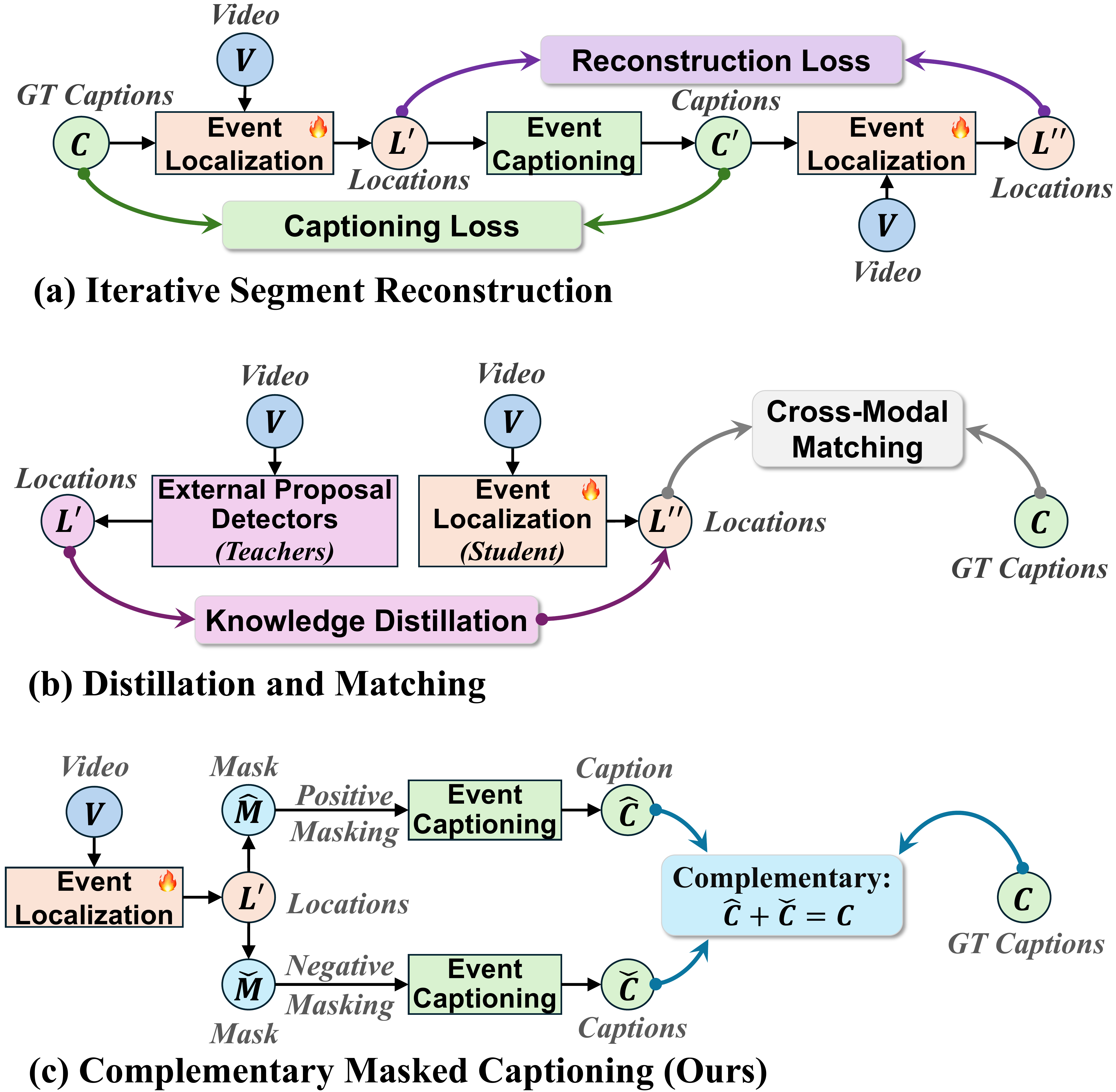}
    \caption{Comparison of our complementary masking paradigm with previous paradigms for event localization.}
    \label{fig:head}
\end{figure}

\section{Introduction}

Dense Video Captioning (DVC) is a challenging task that aims to generate a series of temporally localized captions to describe the various events in a video \cite{krishna2017dense}. 
It extends traditional video captioning by providing a more comprehensive understanding of the video content, making it particularly useful for applications such as video understanding, video summarization, and video search \cite{li2018jointly,wang2021end,yang2023vid2seq}. 
Recently, the Weakly-Supervised Dense Video Captioning (WSDVC) task, which only relies on video-level captions and does not require extensive temporal location annotations for training, has been proposed and appears to be more feasible for practical applications \cite{duan2018weakly}.
However, the lack of supervision on event localization poses a great challenge in accurately locating the events in the video.

To address this problem, existing methods attempt to align event locations and captions using alignment constraints like reconstruction loss or cross-modal matching loss \cite{duan2018weakly,chen2021towards,wu2021weakly,choi2023pws}. 
These approaches primarily fall into two categories: Iterative Segment Reconstruction (Figure \ref{fig:head}(a)) and Distillation and Matching (Figure \ref{fig:head}(b)).
The first type involves a reconstruction cycle where event localization and captioning are interdependent, aiming to minimize reconstruction error \cite{duan2018weakly,chen2021towards,choi2023pws}. 
The second type employs external proposal detectors to guide localization and uses metric learning for cross-modal matching, maximizing semantic similarity between locations and captions \cite{wu2021weakly}.
Despite their promising results, they still suffer from the cumbersome event proposal procedures during both training and inference. 
During the training phase, these methods rely on complex techniques for event localization, such as the use of pre-defined proposals \cite{duan2018weakly,chen2021towards,choi2023pws} or the external pre-trained temporal event localization models \cite{wu2021weakly}. 
During the inference phase, many of them often require a large number of random proposals to be sampled \cite{duan2018weakly,choi2023pws}, which can be computationally expensive. 

Unlike previous methods, in this paper, we propose a novel implicit location-caption alignment paradigm based on complementary masking, which simplifies the complex event proposal and localization process while maintaining effectiveness.
Specifically, our approach involves an dual-mode video captioning module for event captioning and an extra mask generation module for event localization.
We configure our video captioning module to operate in two captioning modes: full video captioning mode and masked video captioning mode.
The first mode can provide global event information (e.g., event count) for the event localization in the second mode, eliminating the need for cumbersome event proposal procedures. 
Besides, as shown in Figure \ref{fig:head}(c), with the mask generation module, we can first predict the temporal location of each event merely based on video and then construct the corresponding location mask. 
After applying a positive mask and its corresponding negative mask (i.e., inverse mask) to the video and performing masked video captioning, we constrain that the captions generated from these two types of masked videos should be complementary (i.e., the two parts of captions constitute the complete video caption).
In this way, even under weak supervision, the event location and event caption can be aligned implicitly. 

In summary, this paper makes the following contributions:
\begin{itemize}[itemsep=1pt,topsep=1pt,parsep=0pt]
    \item We propose a novel implicit location-caption alignment paradigm based on complementary masking, which addresses the problem of unavailable supervision on event localization in the WSDVC task.
    \item We introduce a dual-mode dense video captioning model, which can simplify the process of event localization.
    \item Extensive experiments conducted on the public datasets demonstrate the effectiveness of our method and each of its components.
\end{itemize}

\begin{figure*}[t]
    \centering
    \includegraphics[width=\linewidth]{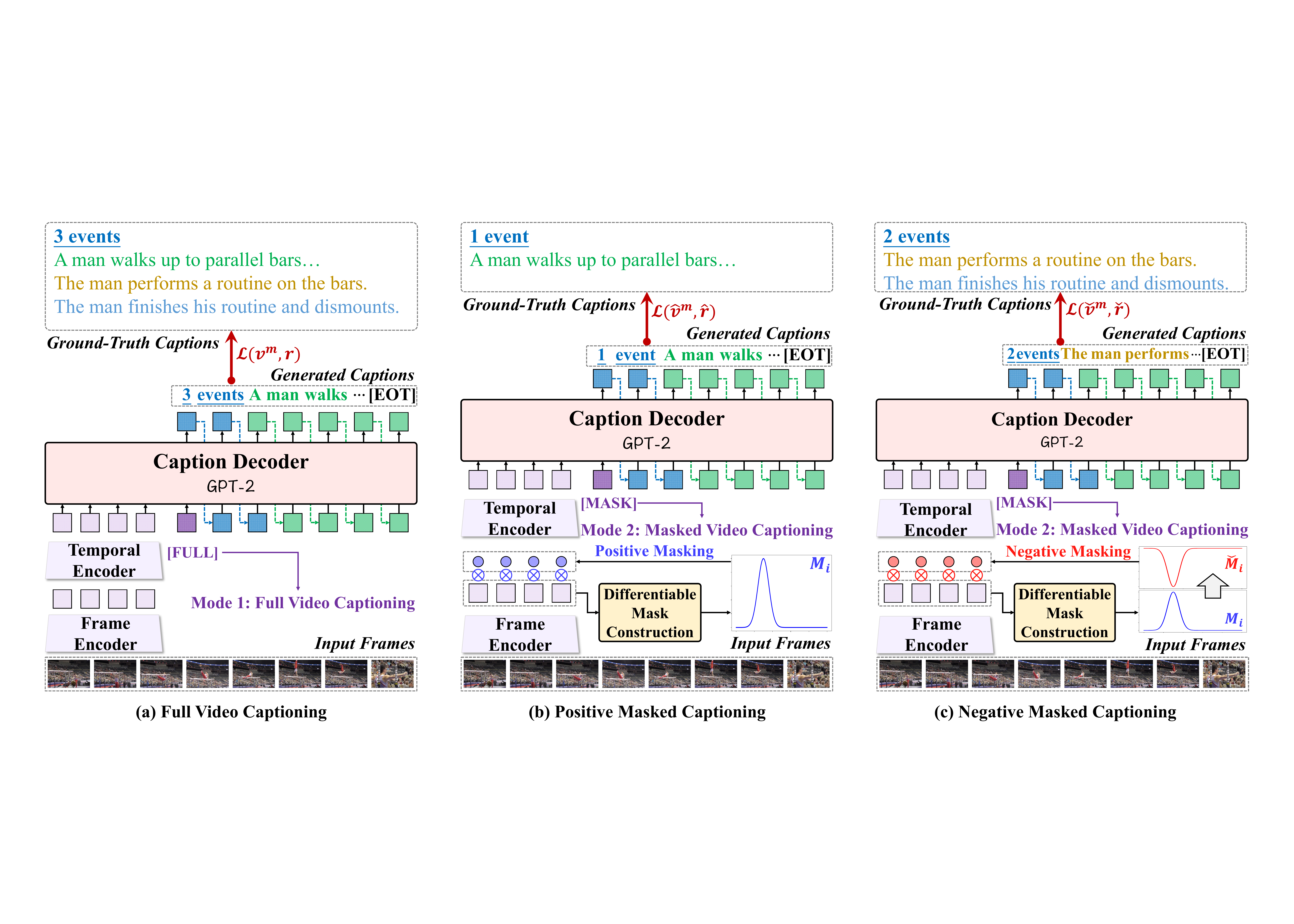}
    \caption{Illustration of our proposed framework, which consists of two main components: a Dense Video Captioning model for event captioning and a Complementary Mask Generation module for event localization.}
    \label{fig:framework}
\end{figure*}

\section{Related Work}
\subsubsection{Dense Video Captioning}
Dense Video Captioning is a challenging multi-task problem that involves event localization \cite{buch2017sst,lin2018bsn,zeng2019graph,zhao2024snippets} and event captioning \cite{gao2017video,seo2022end,nie2022search}.
A lot of existing methods follow the `detect-then-describe' paradigm, which first localizes a set of event proposals and then generates captions for the event proposals \cite{krishna2017dense,wang2018bidirectional,mun2019streamlined,iashin2020multi}.
\citet{krishna2017dense} first introduce the dense video captioning problem and propose an event proposal module followed by a captioning module.
\citet{mun2019streamlined} propose to model temporal dependency across events explicitly and leverages visual and linguistic context from prior events for coherent storytelling.
\citet{iashin2020multi} utilize audio and speech modalities and Transformer architecture to convert multi-modal input data into textual descriptions.
Another line of work removes the explicit event proposing process and jointly performs event localization and captioning for each event \cite{wang2021end,zhu2022end,yang2023vid2seq}.
\citet{wang2021end} formulate the dense caption generation as a set prediction task and feed the enhanced representations of event queries into the localization head and caption head in parallel.
\citet{zhu2022end} propose to solve the dense video captioning task as a single sequence-to-sequence modeling task using a multimodal Transformer. 
\citet{yang2023vid2seq} introduce a single-stage dense event captioning model pretrained on narrated videos and generate event timestamps as special tokens.

\subsubsection{Weakly-Supervised Dense Video Captioning}
Recently, there has been an increased focus on the Weakly-Supervised Dense Video Captioning (WSDVC) setting, which is considered more challenging and practical than the conventional DVC setting \cite{duan2018weakly,wu2021weakly,chen2021towards,choi2023pws}.
\citet{duan2018weakly} first introduce the WSDVC problem and decompose it into the sentence localization and event captioning problems.
Specifically, this paper present a cycle system based on the fixed-point iteration \cite{chidume1987iterative} to train the model.
Several methods follow this cycle training system and enhance the performance by improving the sentence localization and event captioning modules. 
\citet{chen2021towards} propose to use a concept learner as the basis of the sentence localizer, which can be utilized to construct an induced set of concept features to enhance video features and improve the event captioner.
\citet{choi2023pws} further improve the performance by pretraining the event captioning model on an extra video description dataset MSR-VTT \cite{xu2016msr}.
Different from the above methods, \citet{wu2021weakly} adopt the knowledge distilled from relevant tasks to generate high-quality event proposals and build semantic matching between the proposals and sentences.
However, these methods still suffer from the cumbersome event proposal procedures during both training and inference. 
                                        
\section{Proposed Method}

\subsection{Task Definition}
 
In Dense Video Captioning (DVC), a video is represented as $\bm{v} = \{v_i\}_{i=1}^{N_v}$, where $v_i$ denotes the $i$-th frame, and $N_v$ is the total number of frames. The objective is to generate captions $\{S_i\}_{i=1}^{N_s}$ for temporally localized events within the video. Each captioning event $S_i$ encompasses a tuple $(t^s_i, t^e_i, C_i)$, detailing the start time, end time, and the associated caption.

Unlike the DVC task, the Weakly-Supervised Dense Video Captioning (WSDVC) requires the model to generate these temporally localized captioning events without relying on explicit annotations for the start and end times of each event, i.e., $t^s_i$ and $t^e_i$ for each $S_i$ are unavailable during training.
The model should leverage information from the video frames and the provided captions during training to infer the appropriate temporal location and generate accurate, contextually relevant captions during inference.

\subsection{Overview}

Our proposed method integrates two main components: a Dense Video Captioning (DVC) module for event captioning and a Complementary Mask Generation (CMG) module for event localization. 
The DVC module operates in two modes: full captioning mode captures global narratives, while masked captioning mode enhances localization through differentiable masks.
The CMG module predicts temporal locations by utilizing positive and negative masks during the masked captioning mode.
Positive masking focuses on specific event captions, and negative masking handles the remaining context, ensuring alignment under weak supervision. 
Together, these components enable our model to align captions with video locations effectively.

\subsection{Full Video Captions Generation}
Our proposed DVC module leverages a spatial-temporal video encoder and a pretrained language model to generate multiple content-continuous captions for a given video. 
As shown in Figure \ref{fig:framework}(a), the process can be divided into two steps: (1) spatial-temporal video encoding, which captures both spatial and temporal information from the video frames, and (2) prompt-based caption decoding, which fine-tunes the language model using video embeddings to generate contextually relevant captions.

\subsubsection{Spatial-Temporal Video Encoding}
To fully capture visual information, we design a spatial-temporal encoder with a frame-level spatial encoder $E^a$ and a video-level temporal encoder $E^b$.
Given a video $v\in\mathbb{R}^{N_v\times{3}\times{H}\times{W}}$, where $H$ and $W$ represent the height and width of each video frame,the spatial encoder $E^a$ extracts visual embeddings from each frame, resulting in frame-level embeddings $\bm{v}^a\in\mathbb{R}^{N_v\times{d}}$:
\begin{equation}
\bm{v}^a = \{E^a(v_i)\}_{i=1}^{N_v},
\end{equation}
where $d$ is the dimension of the embeddings. 
We formalize $E^a$ using the pretrained CLIP ViT-L/14 model \cite{radford2021learning} and keep the parameters of $E^a$ frozen during training and testing.
Next, the temporal encoder $E^b$ processes these embeddings $\bm{v}^a$ to capture temporal information, generating contextualized video embeddings $\bm{v}^b$:
\begin{equation}
\bm{v}^b = E^b(\bm{v}^a + \theta_p)\in\mathbb{R}^{N_v\times{d}},
\end{equation}
where $\theta_p\in\mathbb{R}^{N_v\times{d}}$ are position embeddings and $E^b$ is a randomly initialized Transformer encoder. 
The output $\bm{v}^b$ encodes frame characteristics and their temporal relationships, essential for generating accurate captions.

\subsubsection{Prompt-Based Caption Decoding}
Inspired by advancements in multimodal language models \cite{li2023blip,zhu2023minigpt,liu2024visual}, we adapt a pretrained GPT-2 \cite{radford2019language} to serve as a prompt-based caption decoder. 
This model processes video embeddings to generate sequential captions for all events in a video.
Given captions $\{C_i\}_{i=1}^{N_S}$, we concatenate a prompt $P$ ``\texttt{[FULL] $N_S$ events:}'' with all captions into a paragraph, tokenized and embedded into a sequence $\bm{r}$:
\begin{equation}
    \bm{r} = \{r_1, \dots, r_{N_r}\} = \text{tokenizer}(\{P, C_1, \dots, C_{N_S}\}),
\end{equation}
where $\texttt{[FULL]}$ signals the model to generate all captions, ``\texttt{$N_S$ events:}'' specifies the number of captions, and $N_r$ is the number of all tokens. 
We use contextualized video embeddings $\bm{v}^b$ as prefix visual tokens, concatenated with caption token embeddings:
\begin{equation}
    Z = \{\bm{v}^b;r_1, \dots, r_{N_r}\}.
\end{equation}
The objective is to minimize the negative log-likelihood of generating caption tokens given video embeddings and previous tokens:
\begin{equation}
    \mathcal{L}(\bm{v}^b,\bm{r}) = -\frac{1}{N_r}\sum_{i=2}^{N_r}\log{p(r_i \mid \bm{v}^b, r_{<i}; \theta_{E,G})},
\end{equation}
where $\theta_{E,G}$ are parameters of the video encoder and caption decoder. 
By combining the video encoding and caption decoding steps, our full-video captioning model effectively generates multiple content-continuous captions for any given video, capturing the various events and their temporal relationships within the video.

\subsection{Captioning-Guided Event Localization}
We introduce a Captioning-Guided Event Localization method to ground the generated captions to the corresponding video segments.
As shown in Figure \ref{fig:framework}(b)\&(c), this method consists of two main steps:
(1) Mask Prediction for Events, predicting the center and width of each caption to create a differentiable Gaussian mask representing its temporal location, 
and (2) Complementary Masked Captioning, applying these masks to video embeddings for training on positive and negative masked captioning tasks.
 
\subsubsection{Mask Prediction for Event}
\begin{figure}[t]
    \centering
    \includegraphics[width=0.85\linewidth]{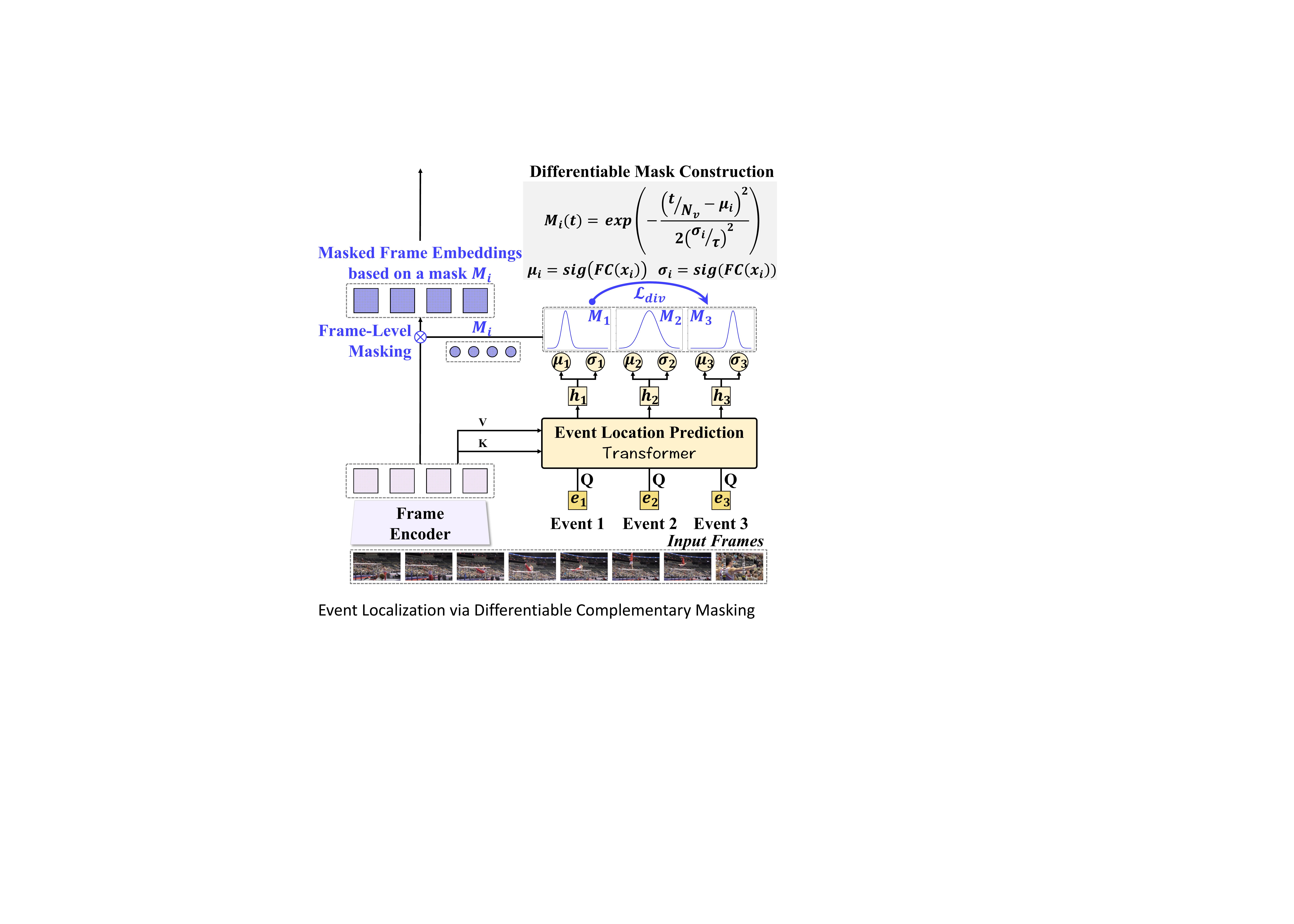}
    \caption{Detailed illustration of Event Location Prediction and Differentiable Mask Construction.}
    \label{fig:mask}    
\end{figure}

As shown in Figure \ref{fig:mask}, the mask prediction component consists of the event caption prediction and the differentiable mask construction processes.
The goal is to generate a differentiable mask for each caption to represent its temporal location within the video and encourage diversity among the masks.

\noindent\textbf{\textit{Event Location Prediction.}}
We first use a randomly initialized learnable embedding as a event embedding $e_i\in\mathbb{R}^d$ for each caption proposal $C_i$. 
Then, for all event embeddings $\bm{e}\in\mathbb{R}^{N_S\times{d}}$ and the frame embeddings $\bm{v}^a\in\mathbb{R}^{N_v\times{d}}$, we use a Transformer decoder to generate the embeddings $\bm{h}$:
\begin{equation}
\bm{h} = \text{TransformerDecoder}(\bm{e}, \bm{v}^a+\theta_p) \in\mathbb{R}^{N_S\times{d}}.
\end{equation}
Next, we utilize two linear layers to predict the center $\mu_i$ and width $\sigma_i$ of the proposal $i$:
\begin{equation}
    \mu_i = \text{Sigmoid}(\text{FC}_1(\bm{h}_i)), \quad \sigma_i=\text{Sigmoid}(\text{FC}_2(\bm{h}_i)).
\end{equation}

\noindent\textit{\textbf{Differentiable Mask Construction.}}
Inspired by previous works \cite{zheng2022weakly,zheng2022weakly2,kim2024gaussian}, we use $\mu_i$ and $\sigma_i$ to generate a Gaussian mask $M_i\in\mathbb{R}^{N_v}$ for each event, representing its temporal location:
\begin{equation}
M_i(t) = \exp\left(-\frac{(t/N_v - \mu_i)^2}{2(\sigma_i/\tau)^2}\right),  \forall t \in \{1, \dots, N_v\},
\end{equation}
where $M_i(t)$ is the mask value at frame $v_t$, and $\tau$ is a hyperparameter that controls the steepness of the Gaussian curve.
Note that other functions generating soft masks can be used as alternatives to the Gaussian mask.

\noindent\textit{\textbf{Diversity Loss.}}
To ensure masks cover different video parts, we introduce a diversity loss based on cosine similarity, serving as a regularization term:
\begin{equation}
\mathcal{L}_{\text{div}} = \frac{1}{N_S(N_S-1)} \sum_{i=1}^{N_S} \sum_{\substack{j=1, j\neq i}}^{N_S} 
\max\left(s(i, j) - \gamma, 0\right)
\end{equation}
where $ s(i, j) = \frac{M_i}{\|M_i\|} \cdot \frac{{M_j}^\top}{\|M_j\|}$ and $\gamma$ is the hyperparameter that controls overlap between masks.

\subsubsection{\textbf{Complementary Masked Captioning}}
Since the supervision for event localization is unavailable, we propose to guide event localization by complementary masked captioning, which mainly consists of two subtasks: positive masked captioning and negative masked captioning.

\noindent\textit{\textbf{Positive Masked Captioning.}}
In this task, we apply the Gaussian mask $M_i$ to the video frame embeddings $\bm{v'}$ in the temporal dimension and input them into the video-level temporal encoder $E^b$:
\begin{equation}
    \bm{\hat{v}}^b_i = E^b(M_i \cdot \bm{v}^a).
\end{equation}
We then concatenate a prompt sentence $\hat{P}$ ``\texttt{[MASK] 1 event:}'' with caption $C_i$ and tokenize it into a sequence $\hat{\bm{r}}_i$:
\begin{equation}
    \hat{\bm{r}}_i = \{\hat{r}_1, \dots, \hat{r}_{N_{\hat{r}}}\} = \text{tokenizer}(\{\hat{P}, C_i\}),
\end{equation}
where \texttt{[MASK]} is a special prompt token that indicates the model to generate part of the captions.

Finally, the optimization objective for Positive Masked Captioning is defined as:
\begin{equation}
\begin{split}
    \mathcal{L}(\bm{\hat{v}}^b,\bm{\hat{r}}) = -\frac{1}{N}\sum_{n=1}^{N_S}\sum_{i=2}^{N_{\hat{r}_n}}\log p(\hat{r}_{n,i}\mid & \bm{\hat{v}}^b_n,  
  \hat{r}_{n,1},\dots,\hat{r}_{n,i-1};\\ &\theta_{E,G,T}),
\end{split}
\end{equation}
where $N$ is the total number of the tokens and $T_{n,i}$ denotes the $i$-th token of the tokenized caption $C_n$ and $\theta_{E,G,T}$ denotes the parameters of the video encoder, the caption decoder and the event localization model, respectively.

\begin{table*}[t]\setlength{\tabcolsep}{3pt}  
    \centering
      \begin{tabular}{llccccccccc}
      \toprule
       & \textbf{Model} & \textbf{Features} & \textbf{SODA} & \textbf{METEOR} & \textbf{CIDEr} & \textbf{ROUGE-L} & \textbf{BLEU-1} & \textbf{BLEU-2} & \textbf{BLEU-3} & \textbf{BLEU-4} \\
       \midrule
      \multirow{4}{*}{\makecell[l]{\textbf{Fully-}\\ \textbf{Supervised}}} & DCE & C3D & $-$ & 5.69 & 12.43 & $-$ & 10.81 & 4.57 & 1.90 & 0.71 \\
       & DVC & C3D & $-$ & 6.93 & 12.61 & $-$ & 12.22 & 5.72 & 2.27 & 0.73 \\
       & PDVC & C3D & 5.26 & 7.50 & 25.87 & $-$ & $-$ & $-$ & $-$ & 1.65 \\
       & Vid2Seq & CLIP & 5.80 & 8.50 & 30.10 & $-$ & $-$ & $-$ & $-$ & $-$ \\
      \midrule
      \multirow{8}{*}{\makecell[l]{\textbf{Weakly-}\\ \textbf{Supervised}}} & WSDEC & C3D & $-$ & 6.30 & 18.77 & 12.55 & 12.41 & 5.50 & 2.62 & 1.27 \\
       & ECG & C3D & $-$ & 7.06 & 14.25 & $-$ & 11.85 & 5.64 & 2.71 & 1.33 \\
       & EC-SL & C3D & $-$ & 7.49 & 21.21 & 13.02 & 13.36 & 5.96 & 2.78 & 1.33 \\
       & PWS-DVC$^{*}$ & C3D & $-$ & 7.28 & 20.59 & 12.71 & $-$ & $-$ & $-$ & 1.35 \\
       \cmidrule{2-11}
       & \textbf{Ours}$^\dagger$ & C3D & 5.20 & 7.36 & 28.00 & 13.22 & 13.66 & 6.58 & 3.29 & 1.77 \\  
       & \textbf{Ours}  & C3D & 5.29 & 7.71 & 30.17 & 13.91 & 14.37 & 7.05 & 3.58 & 1.96 \\
       & \textbf{Ours}$^\dagger$  & CLIP & 6.06 & 8.22 & 30.21 & 14.52 & 14.83 & 7.79 & 3.98 & 2.03 \\  
       & \textbf{Ours} & CLIP & \textbf{6.08} & \textbf{8.48} & \textbf{33.42} & \textbf{14.77} & \textbf{15.36} & \textbf{8.12} & \textbf{4.17} & \textbf{2.26} \\
      \bottomrule
      \end{tabular}%
      \caption{Comparison with existing methods on the ActivityNet Caption dataset. The symbol $\dagger$ indicates the GPT-2 model used in our method is randomly initialized. The symbol $*$ indicates the results without training with extra video captioning datasets, ensuring a fair comparison with other methods. References for the compared methods are: \cite{krishna2017dense,li2018jointly,wang2021end,yang2023vid2seq,duan2018weakly,wu2021weakly,chen2021towards,choi2023pws}} 
    \label{tab:table1}%
\end{table*}% 

\noindent\textit{\textbf{Negative Masked Captioning.}}
This task further explores the alignment between captions and video frames by predicting the remaining captions $\{C_j\}_{j=1,j\neq i}^{N_S}$ using the inverse mask on video embeddings. 
Specifically, we compute the negative Gaussian mask $\check{M}_i$ and inverse masked video embedding $\bm{\check{v}}^b_i$ as:
\begin{equation}
    \check{M_i} = 1 - M_i, \quad \bm{\check{v}}^b_i = E^b(\check{M_i} \cdot \bm{v}^a).
\end{equation}
Next, we concatenate a prompt sentence $\check{P}$ ``\texttt{[MASK] $N_S-1$ events:}'' with captions $\{C_j\}_{j=1,j\neq i}^{N_S}$ and tokenize it into a sequence of tokens $\check{\bm{r}}_i$:
\begin{equation}
    \check{\bm{r}}_i = \{\check{r}_1, \dots, \check{r}_{N_{\check{k}}}\} = \text{tokenizer}(\{\check{P}, \{C_j\}_{j=1,j\neq i}^{N_S}\}).
\end{equation}

Finally, the optimization objective for Negative Masked Captioning is defined as:
\begin{equation}
\begin{aligned}
   \mathcal{L}(\bm{\check{v}}^b,\bm{\check{r}}) = -\frac{1}{N}\sum_{n=1}^{N_S}\sum_{i=2}^{N_{\check{r}_n}}\log p(\check{r}_{n,i}\mid & \bm{\check{v}}^b_n, \check{r}_{n,1},\dots,\check{r}_{n,i-1}; \\ & \theta_{E,G,T}). 
\end{aligned}
\end{equation}

By combining Positive and Negative Masked Captioning tasks, the Complementary Masked Captioning component enables the model to learn a better alignment between event captions and locations.

\subsection{Model Training and Inference}
\subsubsection{Model Training}
Our model training consists of two stages: \textit{captioning} and \textit{localizing}. 
In the captioning stage, we train the DVC module by minimizing $\mathcal{L}(\bm{v}^b,\bm{r})$ to generate multiple captions. 
Next, in the localizing stage, we generate Gaussian masks based on the ground-truth captions in the training set and then compute the positive and negative captioning losses.
We add these losses together to form the optimization objective, which is used to train the whole model:
\begin{equation}
    \mathcal{L} = \mathcal{L}(\bm{\hat{v}}^b,\bm{\hat{r}}) + \mathcal{L}(\bm{\check{v}}^b,\bm{\check{r}}) + \mathcal{L}_{div}.
\end{equation}

\subsubsection{Model Inference}
The inference process consists of three stages: \textit{captioning}, \textit{localizing}, and \textit{refining}. 
In the captioning stage, we use the video encoder and caption decoder to generate initial captions with the video embedding and ``\texttt{[FULL]}'' prompt. 
Unlike training, the model determines the number of captions based on video content. 
In the localizing stage, we predict timestamps and generate Gaussian masks for the coarse captions. 
Finally, in the refining stage, we perform positive captioning with masked video embeddings and ``\texttt{[MASK] 1 events:}'' to re-generate each caption, aiming to enhance caption quality.

\begin{table}[t]\setlength{\tabcolsep}{1pt}  
    \centering
    \small
      \begin{tabular}{lcccccc}
      \toprule
      \multirow{3}{*}{\textbf{Model}} &  \multicolumn{3}{c}{\textbf{YouCook2}} & \multicolumn{3}{c}{\textbf{ViTT}} \\
      \cmidrule(lr){2-4}\cmidrule(lr){5-7} & SODA & METEOR  & CIDEr & SODA & METEOR &  CIDEr \\
      \midrule
      WSDEC$^\ddagger$ & 2.11 & 1.47 & 8.43 & 4.13 & 1.95 & 10.31  \\
      PWS-DVC$^\ddagger$ & 3.14 & 2.48 & 9.81 & 6.11 & 2.36 & 12.53 \\
      \midrule
      Ours  & \textbf{3.60} & \textbf{4.77} & \textbf{13.38} & \textbf{8.54} & \textbf{4.21} & \textbf{17.28} \\
      \bottomrule
      \end{tabular}%
    \caption{Comparison with existing methods on the YouCook2 and ViTT datasets. $\ddagger$ means we reimplement and rerun the baseline methods on the two new datasets.}
    \label{tab:vitt}
\end{table}% 

\section{Experiments}

\subsection{Experiment Setup}
\subsubsection{Datasets}
We evaluate our proposed method and baseline methods on the \textbf{ActivityNet Captions} dataset.
The dataset connects videos to a series of temporally annotated sentence descriptions.
Besides, we also conduct experiments on \textbf{ViTT} \cite{huang2020multimodal} and \textbf{YouCook2} \cite{zhou2018towards}, which are two DVC datasets and have never been used for the evaluation of WSDVC methods. 

\begin{table*}[t]\setlength{\tabcolsep}{3pt}   
    \centering
      \begin{tabular}{lccccccccc}
      \toprule
      \textbf{Setting} & \textbf{SODA} & \textbf{METEOR} & \textbf{CIDEr} & \textbf{ROUGE-L} & \textbf{BLEU-1} & \textbf{BLEU-2} & \textbf{BLEU-3} & \textbf{BLEU-4} \\
      \midrule
      \textbf{Full} & \textbf{6.08} & \textbf{8.48} & \textbf{33.42} & \textbf{14.77} & \textbf{15.36} & \textbf{8.12} & \textbf{4.17} & \textbf{2.26} \\
      \midrule
      $-$ Temporal video encoder & 5.88 & 8.33 & 32.64 & 14.40 & 14.98 & 7.88 & 4.05 & 2.18 \\
      $-$ \texttt{[FULL]} or \texttt{[MASK]} prompt & 5.96 & 8.43 & 33.04 & 14.53 & 15.08 & 7.95 & 4.08 & 2.16 \\
      $-$ ``\texttt{$N$ events}'' prompt & 5.82 & 8.30 & 32.36 & 14.17 & 14.87 & 7.95 & 3.98 & 2.15 \\
      $-$ Inference refinement & 5.76 & 8.12 & 31.29 & 14.21 & 14.71 & 7.52 & 3.82 & 2.05 \\
      \midrule
      $-$ Gaussian mask, $+$ Hard binary mask & 3.89 & 6.52 & 16.96 & 11.24 & 11.64 & 5.01 & 1.96 & 0.79 \\
      $-$ Gaussian mask, $+$ Sigmoid mask & 5.98 & 7.95 & 27.79 & 14.03 & 14.68 & 7.41 & 3.58 & 1.85 \\
      $-$ Gaussian mask, $+$ Cauchy mask & 6.02 & 8.32 & 32.22 & 14.50 & 15.15 & 7.98 & 4.07 & 2.21 \\
      \midrule
      $-$ Positive masked captioning & 4.08 & 7.10 & 20.22 & 12.43 & 12.48 & 5.71 & 2.31 & 1.14  \\
      $-$ Negative masked captioning & 5.92 & 8.13 & 30.29 & 14.37 & 15.00 & 7.71 & 3.88 & 1.91 \\
      $-$ Diversity loss $\mathcal{L}_{div}$ & 5.73 & 8.22 & 30.93 & 14.48 & 15.14 & 7.79 & 3.84 & 2.01 \\      
      \bottomrule
      \end{tabular}%
    \caption{Ablation study of our proposed method. The symbol `$-$' means removing the component.}
    \label{tab:table2}%
\end{table*}% 
 
\subsubsection{Evaluation Metrics} 
To make a fair comparison with previous methods, we use the evaluation tool provided by the 2018 ActivityNet Captions Challenge, which measures the capability to localize and describe events.
To clarify, we calculate the METEOR \cite{banerjee2005meteor}, CIDEr \cite{vedantam2015cider}, ROUGE-L \cite{lin2004rouge}, and BLEU-N \cite{papineni2002bleu} scores for the generated captions by comparing them to the reference captions. 
Moreover, we also adopt the recently proposed SODA metric \cite{fujita2020soda} to perform an overall evaluation of our proposed method.
 
\subsubsection{Implementation Details} 
We set the number of transformer blocks in the video-level temporal encoder and cross-modal localizer to 6 and 1, respectively.
The number of attention heads, dimension of hidden states, and feed-forward layers are set to 12, 768, and 2, 048 in all transformer blocks, respectively.
We utilize the Distilled-GPT2 model for the construction of our caption decoder model.
For the training of the model, we adopt the AdamW \cite{loshchilov2017decoupled} optimizer with an initial learning rate of 1e-4 with a warmup rate of 0.1.
We train the model for 10 epochs for the captioning stage and 10 epochs for the localizing stage on 8 Tesla V100 GPUs with a batch size of 8.

\begin{table}[t]\setlength{\tabcolsep}{1pt}  
    \centering
      \begin{tabular}{lcccccc}
      \toprule
      \textbf{Setting} & \textbf{Backbone} & \textbf{Size} (M) & \textbf{SODA} & \textbf{CIDEr} \\
      \midrule
      PWS-DVC$^{*\dagger}$ & vanilla Transformer  & $\sim$ 59 & $-$ & 20.59 \\
      Vid2Seq & T5-Base & 264.48 & 5.80 & 30.10 \\
      Ours$^\dagger$(C3D) & Distilled-GPT2 & 62.10 & 5.20 & 28.00 \\
      Ours$^\dagger$ & Distilled-GPT2 & 62.10 & 6.06 & 30.21 \\
      Ours & Distilled-GPT2 & 62.10 & \textbf{6.08} & \textbf{33.42} \\
      Ours$^\dagger$ & GPT2-Base & 104.62 & 5.74 & 26.92 \\
      Ours & GPT2-Base & 104.62 & 6.00 & 32.77 \\
      \bottomrule
      \end{tabular}%
    \caption{Comparison of the size and performance of different models. $\dagger$ indicates that the backbone is not pretrained.}
    \label{tab:table3}%
\end{table}% 
 
\subsection{Comparison with Existing Methods}
Table \ref{tab:table1} presents a comparison of our proposed method with existing fully-supervised and weakly-supervised methods on the ActivityNet Caption dataset. 
As shown in Table \ref{tab:table1}, our method outperforms all existing weakly-supervised methods across all evaluation metrics. 
Specifically, our model trained with CLIP features achieves the highest METEOR, CIDEr, ROUGE-L, BLEU-N, and SODA scores, demonstrating its effectiveness in generating accurate and contextually relevant captions. 
Besides, our model trained with C3D features also shows strong performance, outperforming other weakly-supervised methods in most metrics. 
Notably, even the version of our method with a randomly initialized GPT-2 model (denoted by $\dagger$) achieves competitive results, indicating the robustness of our approach.
Moreover, as shown in Table \ref{tab:vitt}, our method also outperform existing methods on the YouCook2 and ViTT datasets.
In summary, our proposed method demonstrates excellent performance in the weakly-supervised dense video captioning task, outperforming existing weakly-supervised methods. 

\subsection{Ablation Study}
In this section, we conduct an ablation study to investigate the contribution of each component in our proposed method. 
The results are shown in Table \ref{tab:table2}.
More details are shown in Supplementary Material.

\begin{figure*}[t]
    \centering
    \subfloat[METEOR and CIDEr scores.]{
        \includegraphics[width=0.238\linewidth]{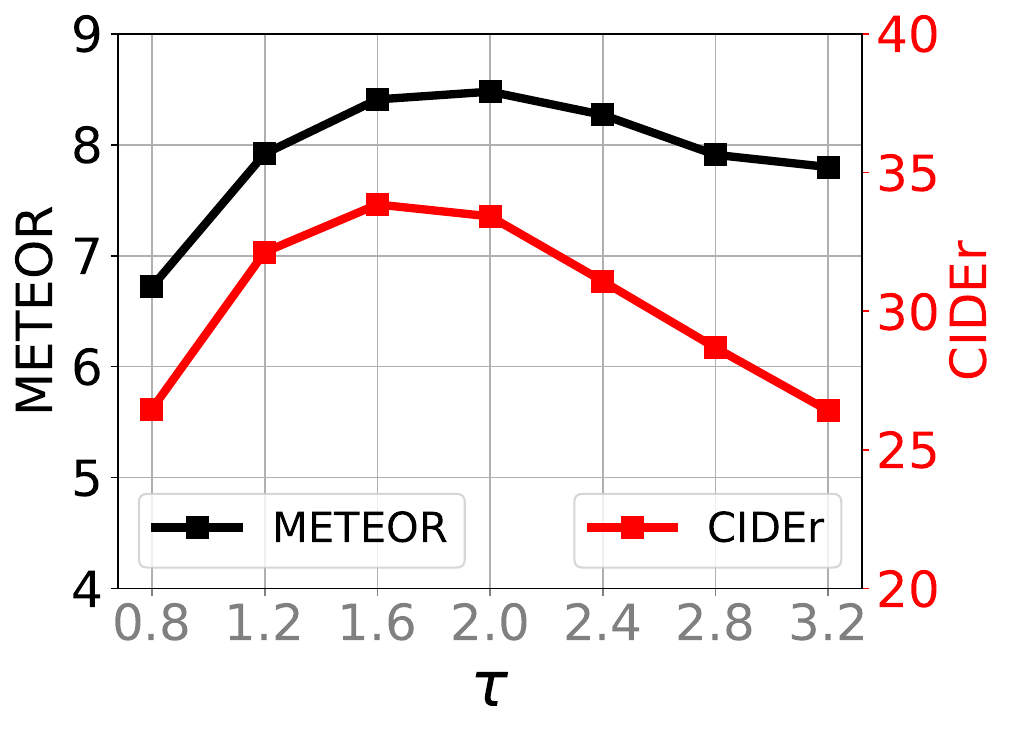}}
    \subfloat[BLEU-4 and SODA scores.]{
        \includegraphics[width=0.242\linewidth]{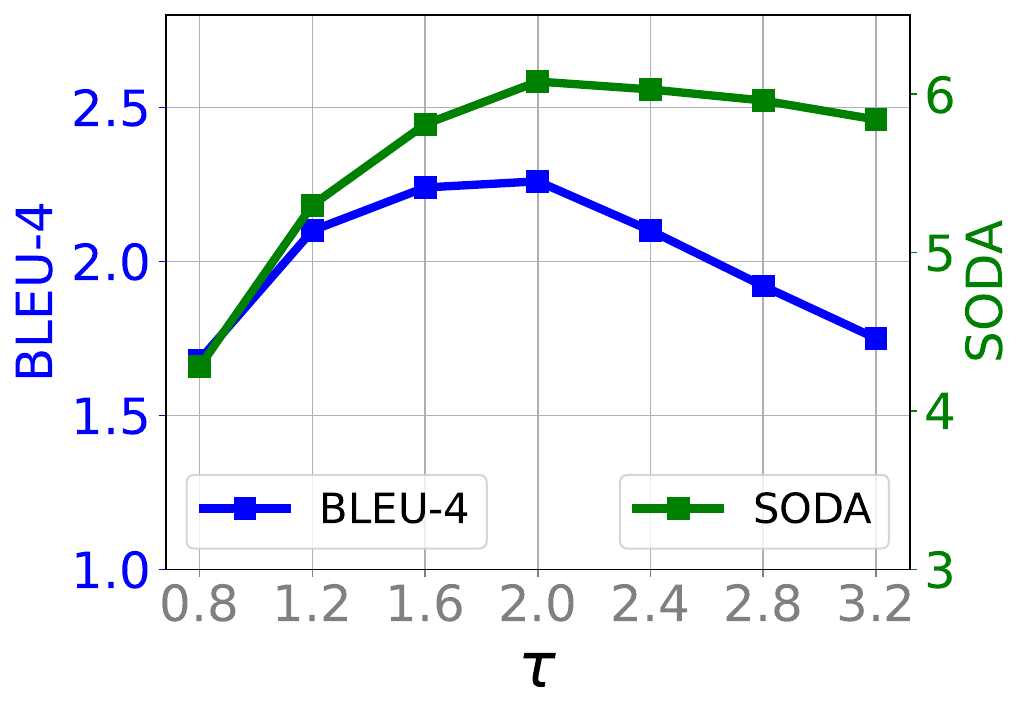}}
    \subfloat[METEOR and CIDEr scores.]{
        \includegraphics[width=0.248\linewidth]{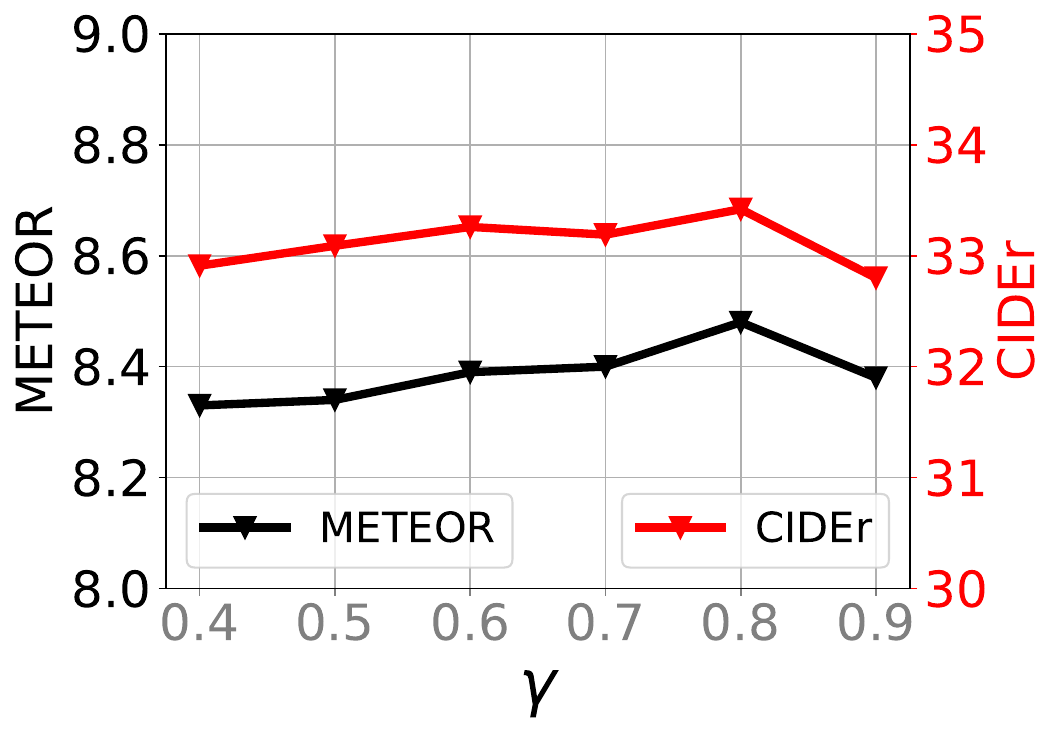}}
    \subfloat[BLEU-4 and SODA scores.]{
        \includegraphics[width=0.252\linewidth]{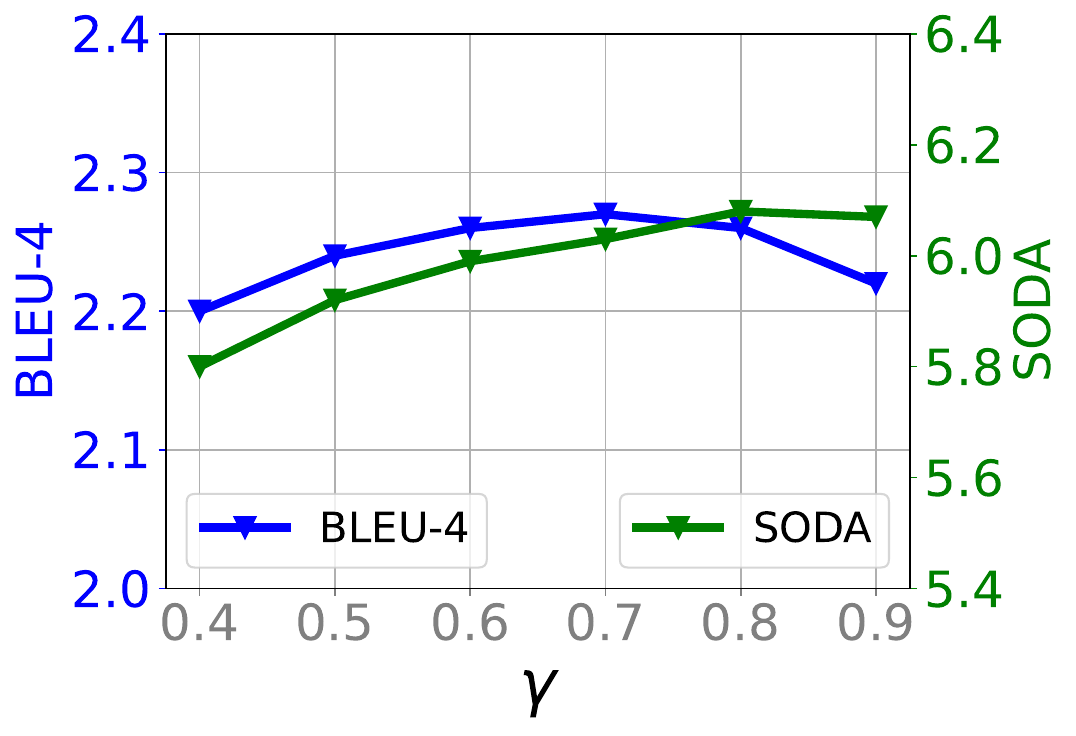}}
    \caption{Impact of $\tau$ in the Gaussian mask construction (a-b) and impact of $\gamma$ in the diversity loss (c-d).}
    \label{fig:hp}
\end{figure*}

\subsubsection{Effect of the model design} 
As shown in Table \ref{tab:table2}, our full model achieves the best performance across all metrics, demonstrating the effectiveness of our proposed method.
When we remove the temporal video encoder, the performance in all metrics decreases slightly. 
We can see that using CLIP features leads to better performance compared to using C3D features. 
This indicates that the choice of spatial frame encoder has a significant impact on the model's performance, and using a more powerful encoder like CLIP can further improve the quality of generated captions.
Additionally, we observe a drop in performance when we remove the \texttt{[FULL]} or \texttt{[MASK]} prompt, which suggests that these prompts are important for guiding the model to generate appropriate captions. 
Similarly, removing the ``\texttt{$N$ events}'' prompt leads to a decrease in performance, indicating its importance in informing the model about the number of events in the video.
Moreover, removing the refining stage in the model inference process could also leads to a noticeable performance decline.
Overall, the results indicates that each component plays crucial roles in capturing video information and improving the quality of generated captions. 

\subsubsection{Effect of the mask construction methods}
We also investigate the effect of different mask construction methods. 
The Gaussian mask used in our full model achieves the best performance. 
When we replace the Gaussian mask with a hard binary mask (where inside the predicted scope is 1, otherwise 0), the performance drops significantly, especially in terms of SODA, METEOR, CIDEr, and BLEU scores. 
This result can be attributed to the fact that the hard binary mask is non-differentiable, which prevents the optimization of the event proposal generation process.
Replacing the Gaussian mask with a Sigmoid mask \cite{duan2018weakly} or a Cauchy mask (based on Cauchy Probability density function) also leads to a decrease in performance, although the decrease is less severe than that with the hard binary mask. 

\subsubsection{Effect of the loss functions}
Finally, we examine the effect of the captioning loss functions. 
Removing the positive masked captioning task results in a significant decrease in performance across all metrics. 
Removing the negative masked captioning task also leads to a decrease in performance across all metrics, although the decrease is less severe than that caused by removing the positive captioning task. 
Similarly, removing the diversity loss $\mathcal{L}_{div}$ leads to a decrease in performance. 
This indicates that the diversity loss is important for encouraging the model to generate diverse captions, which can cover different aspects of the video content.
In summary, our ablation study demonstrates that all components of our proposed method, including the captioning loss functions, contribute to its strong performance in the weakly-supervised dense video captioning task.

\subsection{Model Analysis}
\subsubsection{Impact of the model size}
Table \ref{tab:table3} presents the comparison of the model size and performance of different models. 
Our method with a distilled GPT-2 \cite{radford2019language} backbone, even without pretraining, achieves a SODA score of 6.06 and a CIDEr score of 30.21, outperforming the PWS-DVC method which uses a vanilla Transformer \cite{vaswani2017attention} backbone and has a similar model size. 
When equipped with the pretrained distilled GPT-2 backbone, our method achieves even better performance, with a SODA score of 6.08 and a CIDEr score of 33.42. 
This performance is competitive with the Vid2Seq method, which is trained under the fully-supervised settings and uses a larger T5-Base \cite{raffel2020exploring} backbone.
We also explore the effect of using a larger GPT-2 Base backbone in our method. 
The performance improves slightly compared to using the distilled GPT-2 backbone.
However, the improvement is not proportional to the increase in model size, which may be due to overfitting caused by limited training data.
In summary, our proposed method demonstrates strong performance on the WSDVC task, even with a relatively smaller model size.

\begin{figure}[t]
    \centering
    \includegraphics[width=\linewidth]{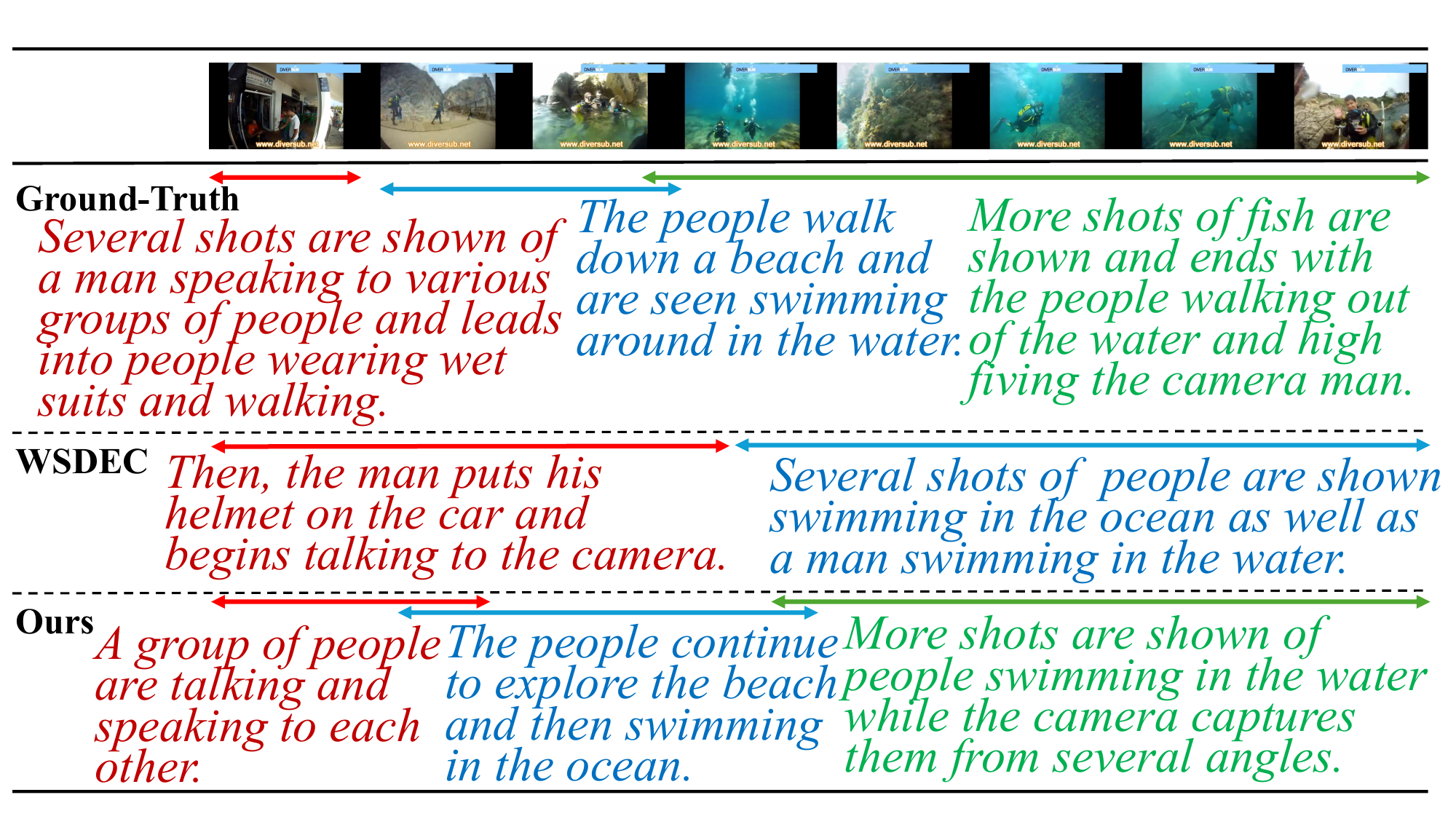}
    \caption{A Qualitative Example from Activity Caption.}
    \label{fig:case}
\end{figure}

\subsubsection{\textbf{Impact of the hyperparameters}}
We examine the impact of values of $\tau$ on the performance of our method. 
The results are shown in Figure \ref{fig:hp}(a-b).
We observe that all the scores increase as $\tau$ increases from 0.8 to 2.0, and then start to decrease when $\tau$ is larger than 2.0. 
Then, to investigate the impact of the $\gamma$ value on our model's performance, we conduct experiments with different $\gamma$ values ranging from 0.4 to 0.9. 
The results are shown in Figure \ref{fig:hp}(c-d).
We observe that most scores generally increase as $\gamma$ increases from 0.4 to 0.8, and then slightly decrease when $\gamma$ reaches 0.9. 
The results show that both the value of $\tau$ and $\gamma$ have the moderate impact on the performance of our model.

\subsubsection{Case study}
Figure \ref{fig:case} compares the predictions of our model with the ground-truth annotations and WSDEC method on an example from the ActivityNet Caption dataset. 
As shown in the cases, our method accurately detects most of the scenes and activities in the video. 
Moreover, our method can roughly predict the number and the location of the events.
In summary, the case study demonstrates the effectiveness of our proposed method for the WSDVC task.

\section{Conclusion}
In this paper, we propose a novel WSDVC method that effectively addresses the problem of unavailable supervision on event localization by implicitly aligning event location with event captions via complementary masking, which simplifies the complex event proposal and localization process while maintaining effectiveness.
Extensive experiments on the public datasets validate the effectiveness of our method.

\section{Acknowledgments}
This work is supported by the National Natural Science Foundation of China under Grants Nos. 61972192, 62172208, 61906085.
This work is partially supported by Collaborative Innovation Center of Novel Software Technology and Industrialization.
This work is supported by the Fundamental Research Funds for the Central Universities under Grant No. 14380001.

\bibliography{paper}

\cleardoublepage
\section{Supplementary Materials}

\subsection{Alternative Mask Construction Methods}

In Table 3 of the paper, we present an ablation study comparing the Gaussian masking method with alternative mask construction methods, including Hard binary, Sigmoid, and Cauchy masks. 
Here, we provide detailed descriptions of these methods. 
Given the predicted center $\mu_i$ and width $\sigma_i$ of proposal $i$, the mask $M_i$ can be calculated as follows:
\subsubsection{Hard binary mask:}
\begin{equation}
\begin{split}
M_i(t) = &
\begin{cases}
    1, & \text{if } \mu_i - \frac{\sigma_i}{2} \leqslant t \leqslant \mu_i + \frac{\sigma_i}{2} \\
    0, & \text{otherwise} \\
\end{cases},  \\
&\forall t \in \{1, \dots, N_v\}.
\end{split}
\end{equation}

\subsubsection{Sigmoid mask:}
\begin{equation}
    M^L_i(t) = \text{Sigmoid}(\tau(t-(\mu_i - \frac{\sigma_i}{2}))),
\end{equation}
\begin{equation}
    M^R_i(t) = \text{Sigmoid}(\tau((\mu_i + \frac{\sigma_i}{2}) - t)),
\end{equation}
\begin{equation}
    M_i(t) = M^L_i(t) \times M^R_i(t), \quad \forall t \in \{1, \dots, N_v\}.
\end{equation}

\subsubsection{Cauchy mask:}
\begin{figure}[t]
    \centering
    \includegraphics[width=\linewidth]{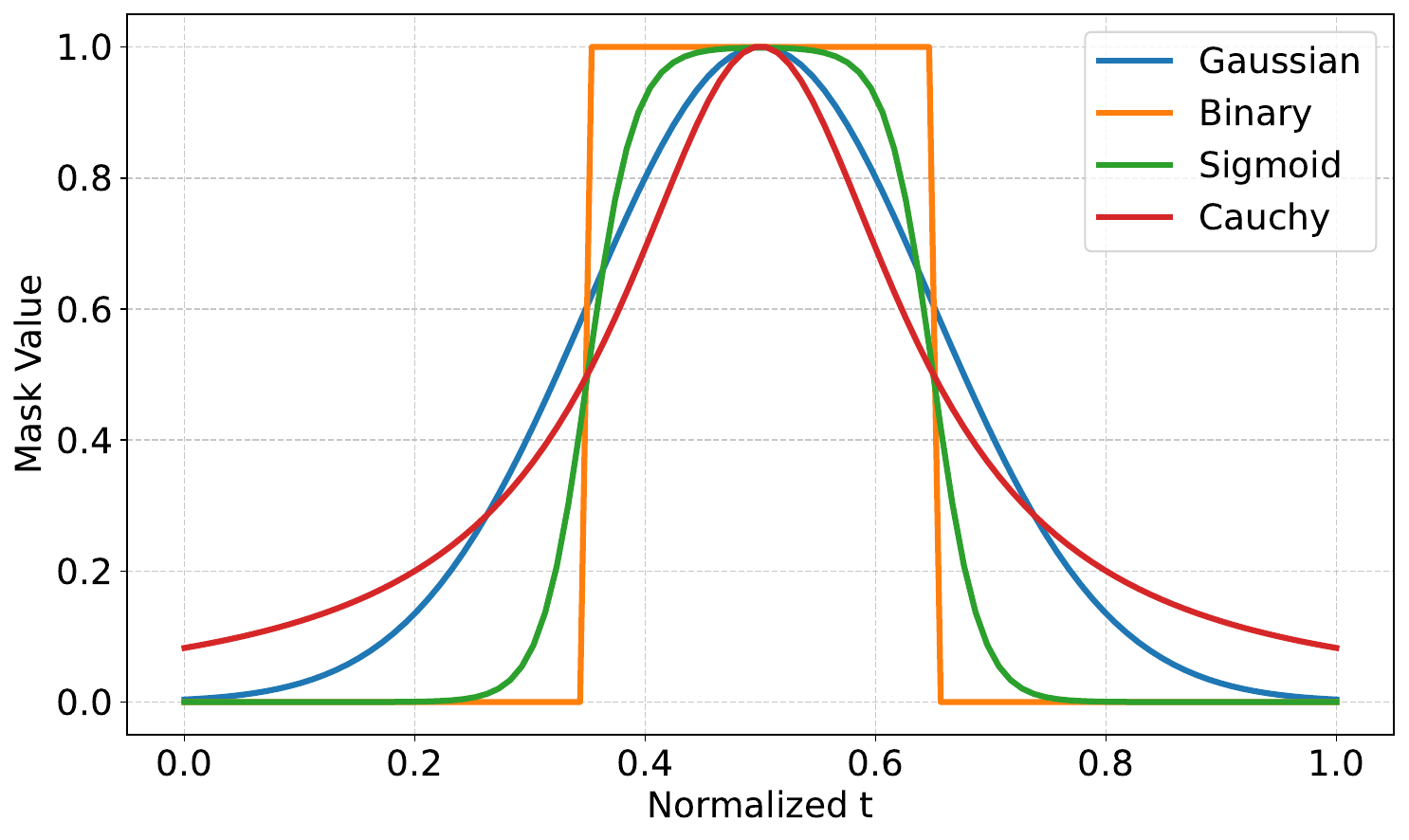}
    \caption{Comparison of different masks with $\mu = 0.5$ and $\sigma = 0.3$.}
    \label{fig:mask2}    
\end{figure}

\begin{equation}
    M'_i(t) = \frac{1}{\pi}\cdot\frac{\sigma_i/\tau}{(t-\mu_i)^2 + (\sigma_i/\tau)^2}
\end{equation}
\begin{equation}
    M_i(t) = \frac{M'_i(t)}{max(M'_i)}, \quad \forall t \in \{1, \dots, N_v\},
\end{equation}
where $\tau$ is the hyperparameter that controls the steepness of the curve. An example of the masks are show in Figure \ref{fig:mask2}.

\subsection{Details of Model Training and Inference}

This section provides an example to explain the details of model training and inference.

\subsubsection{Model Training}

The training process consists of two stages: \textbf{captioning} and \textbf{localizing}.

Suppose we have a training video with 3 captions: ``Sentence A.'', ``Sentence B.'', and ``Sentence C.'' 

\begin{itemize}
    \item \textbf{Captioning Stage}: 
    \begin{itemize}
        \item We train the DVC module by minimizing the loss $\mathcal{L}(\bm{v}^b,\bm{r})$ to generate multiple captions.
        \item Input captions are: ``\texttt{[FULL]} Sentence A. Sentence B. Sentence C.''
        \item These full captions are used to train the video encoder and caption decoder in an autoregressive manner with teacher forcing.
    \end{itemize}

    \item \textbf{Localizing Stage}:
    \begin{itemize}
        \item Gaussian masks are generated based on ground-truth captions.
        \item Positive captions $\bm{\hat{r}}$: ``\texttt{[MASK] 1 events:} Sentence A.'', ``\texttt{[MASK] 1 events:} Sentence B.'', ``\texttt{[MASK] 1 events:} Sentence C.''
        \item Negative captions $\bm{\check{r}}$: ``\texttt{[MASK] 2 events:} Sentence A. Sentence B.'', ``\texttt{[MASK] 2 events:} Sentence A. Sentence C.'', ``\texttt{[MASK] 2 events:} Sentence B. Sentence C.''
        \item The optimization objective for the whole model is:
        \begin{equation}
            \mathcal{L} = \mathcal{L}(\bm{\hat{v}}^b,\bm{\hat{r}}) + \mathcal{L}(\bm{\check{v}}^b,\bm{\check{r}}) + \mathcal{L}_{div}.
        \end{equation}
    \end{itemize}
\end{itemize}

\begin{table}[t]\setlength{\tabcolsep}{4 pt}  
    \centering
      \begin{tabular}{lrrrrrr}
      \toprule
      \textbf{Dataset} & \textbf{Train} & \textbf{Val} & \textbf{Test} & \textbf{Dur} \footnotesize(mins) & \textbf{Sents} \\
      \midrule
      ActivityNet & 10,024 & 4, 926 & 5,044 & 2.0 & 3.7 \\
      ViTT & 4915 & 2487 & 2471 & 4.2 & 7.1 \\
      YouCook2 & 1333 & 457 & 210 & 5.3 & 7.7 \\
      \bottomrule
      \end{tabular}%
    \caption{Statistics of the datasets. `\textbf{Train/Val/Test}' means the number of the videos in the training/validation/testing set. `\textbf{Dur}' means the average duration of each video in the dataset. '\textbf{Sents}' means the average number of annotated sentences in each video.}
    \label{tab:statistics}%
\end{table}% 

\begin{table*}[t]\setlength{\tabcolsep}{4pt}  
    \centering
      \begin{tabular}{crrrrrrrrr}
      \toprule
      \textbf{Backbone} & \textbf{Size} (M) & \textbf{SODA} & \textbf{METEOR} & \textbf{CIDEr} & \textbf{ROUGE-L} & \textbf{BLEU-1} & \textbf{BLEU-2} & \textbf{BLEU-3} & \textbf{BLEU-4} \\
      \midrule
      Distilled-GPT2 & 62.10 & \textbf{6.08} & \textbf{8.48} & \textbf{33.42} & \textbf{14.77} & \textbf{15.36} & \textbf{8.12} & \textbf{4.17} & \textbf{2.26} \\
      GPT2 & 163.89 & 6.00 & 8.13 & 32.77 & 13.81 & 14.58 & 7.81 & 3.77 & 2.14 \\
      T5-Small & 121.37 & 4.72 & 5.98 & 20.28 & 11.60 & 12.43 & 4.78 & 1.40 & 0.33 \\
      T5-Base & 262.36 & 4.76 & 6.08 & 21.23 & 11.23 & 12.34 & 4.76 & 1.42 & 0.43 \\
      \bottomrule
      \end{tabular}%
    \caption{Comparison of the size and performance of different models. $\dagger$ indicates that the backbone is not pretrained.}
    \label{tab:backbone}%
\end{table*}% 

\subsubsection{Model Inference}

The inference process includes three stages: \textbf{captioning}, \textbf{localizing}, and \textbf{refining}.

\begin{itemize}
    \item \textbf{Captioning Stage}:
    \begin{itemize}
        \item The video encoder and caption decoder generate initial captions using the ``\texttt{[FULL]}'' prompt.
        \item For an unlabeled video, generate captions: ``\texttt{[FULL] N events:} Sentence 1. Sentence 2. \dots Sentence N.''
        \item Unlike training, the model determines the number of captions \( N \) based on video content.
    \end{itemize}

    \item \textbf{Localizing Stage}:
    \begin{itemize}
        \item Predict $N$ timestamps and generate Gaussian masks for captions ``Sentence 1.'', ``Sentence 2.'', \dots, ``Sentence N.''
    \end{itemize}

    \item \textbf{Refining Stage}:
    \begin{itemize}
        \item Perform positive captioning with masked video embeddings and ``\texttt{[MASK] 1 events:}'' to regenerate each caption, enhancing quality.
        \item Generated refined content: ``\texttt{[MASK] 1 events:} Sentence \(\bar{1}\).'', ``\texttt{[MASK] 1 events:} Sentence \(\bar{2}\).'', ``\texttt{[MASK] 1 events:} Sentence \(\bar{N}\).''
    \end{itemize}
\end{itemize}

\subsection{Details of Experimental Settings}
\subsubsection{Datasets}
We use the ActivityNet Captions \cite{chen2019activitynet}, ViTT \cite{zhou2018towards}, and YouCook2 \cite{huang2020multimodal} dataset to evaluate our method and baseline methods.
These dataset connects videos to a series of temporally annotated sentence descriptions, covering various themes.
The statistics are shown in Table \ref{tab:statistics}.

\subsubsection{Evaluation Metrics}
We use the evaluation tool provided by the 2018 ActivityNet Captions Challenge, which measures the capability to localize and describe events.
To clarify, we calculate the METEOR \cite{banerjee2005meteor}, CIDEr \cite{vedantam2015cider}, ROUGE-L \cite{lin2004rouge}, and BLEU-N \cite{papineni2002bleu} scores for the generated captions by comparing them to the reference captions. 
We use tIoU (temporal Intersection over Union) thresholds of 0.3, 0.5, 0.7, and 0.9 for this evaluation. 
The final scores are determined by averaging the results across these different thresholds.
Moreover, we also adopt the recently proposed SODA metric \cite{fujita2020soda} to perform an overall evaluation of our proposed method.

\subsection{Impact of the Backbone}
In this paper, we utilize the Distilled-GPT2 model for the construction of our caption decoder model.
We also validate the impact of using different pretrained models as the caption decoder.
The results are shown in Table \ref{tab:backbone}.
Note that the reported model size includes all parameters, which differ from the numbers reported in the paper.
Our model with Distilled-GPT2 model as backbone achieves the best performance.

\subsection{Impact of the Number of Video Frames}
\begin{figure}[t]
    \centering
    \subfloat[METEOR and CIDEr scores.]{
        \includegraphics[width=0.495\linewidth]{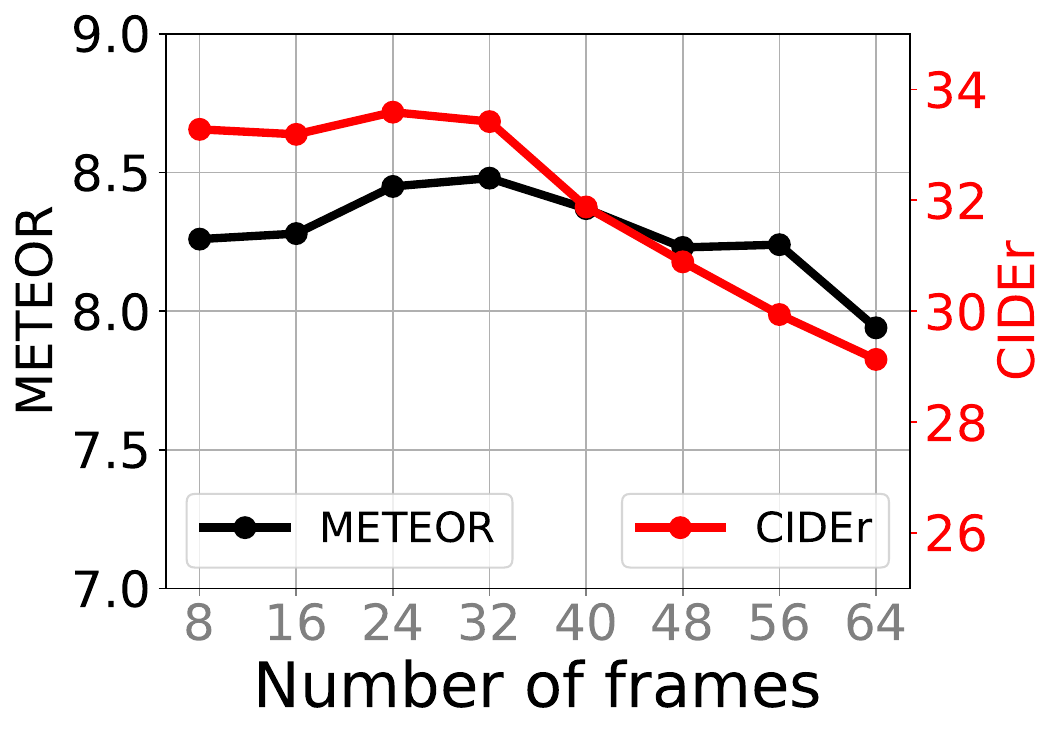}}
    \subfloat[BLEU-4 and SODA scores.]{
        \includegraphics[width=0.5\linewidth]{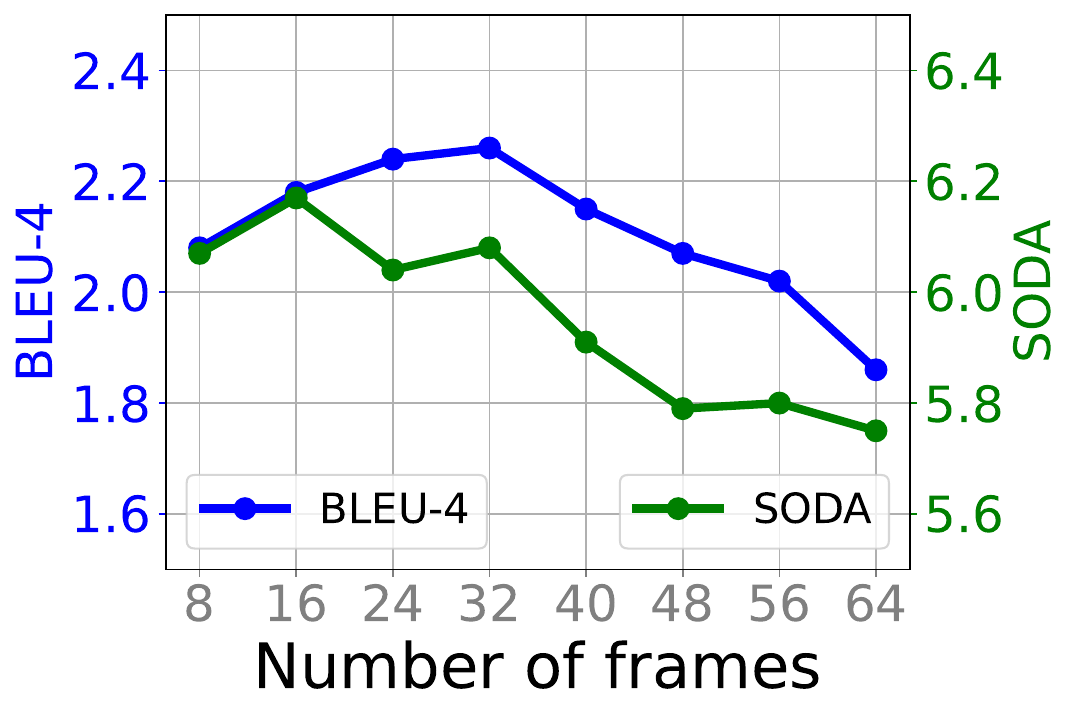}}
    \caption{Impact of the number of video frames.}
    \label{fig:num_frames}
\end{figure}
We analyze the impact of varying the number of video frames on the performance of our proposed method. 
We conduct experiments with different numbers of frames, ranging from 8 to 64. 
The results are shown in Figure \ref{fig:num_frames}.
The first figure displays the METEOR and CIDEr scores as the number of frames increases. 
It can be observed that all the scores initially increase as the number of frames increases, reaching their peak values at 24 and 32 frames, respectively. 
However, as the number of frames continues to increase beyond these points, the performance starts to decrease. 
This suggests that using an appropriate number of frames is important for capturing sufficient temporal information while avoiding the negative effects of excessive frames, such as increased computational complexity and potential noise.
In summary, our analysis indicates that using an appropriate number of video frames is crucial for achieving optimal performance in the WSDVC task. 

\begin{table}[t]\setlength{\tabcolsep}{1.2pt}  
    \centering
      \begin{tabular}{llcccc}
      \toprule
      & \textbf{Method} &\textbf{Features} & \textbf{R@Avg} & \textbf{P@Avg} & \textbf{F1} \\
      \midrule
      \multirow{3}{*}{\makecell[l]{\textbf{Fully-}\\ \textbf{Supervised}}} & SDVC & C3D & 55.58 & 57.57 & 56.56 \\
      & PDVC & TSN & 55.42 & 58.07 & \underline{56.71} \\
      & Vid2Seq & CLIP & \underline{59.0} & \underline{60.3} & $-$ \\
      \midrule
      \multirow{4}{*}{\makecell[l]{\textbf{Weakly-}\\ \textbf{Supervised}}} & WSDEC$^{\dagger}$ & C3D &  29.57 & 59.33 & 39.18 \\
      & PWD-DVC$^{\dagger}$ & C3D & 40.85 & 55.82 & 47.09 \\
      & Ours & C3D & 53.09 & \textbf{59.20} & 55.98 \\
      & Ours & CLIP & \textbf{53.72} & 58.92 & \textbf{56.20} \\
      \bottomrule
      \end{tabular}%
    \caption{Comparison of the event localization performance across various fully-supervised and weakly-supervised models. $\dagger$ indicates that results are obtained by re-running methods in this setting, as the original papers didn’t provide them.}
    \label{tab:loc}%
\end{table}% 

\subsection{Event localization performance}
 We further validate the temporal event localization performance by comparing our method with fully-supervised and weakly supervised methods on the ActivityNet Captions validation set. 
 We adopt the experimental setting used in the work of SDVC as the unified setting.
 As shown in the table \ref{tab:loc}, the performance of our method is significantly better than that of the two weakly-supervised methods and is only slightly worse than that of the supervised DVC methods, which validate the effectiveness of our method on localizing events in the videos.

 \begin{table}[t]\setlength{\tabcolsep}{5pt}  
    \centering
      \begin{tabular}{lccccc}
      \toprule
      \textbf{Setting} & \textbf{0 $\sim$ 0.1} & \textbf{0.1 $\sim$ 0.3} & \textbf{0.3 $\sim$ 0.5} & \textbf{0.5 $\sim$ 1} \\
      \midrule
      Hard & 6.75 & 6.04 & 5.42 & 4.67 \\
      Gaussian & \textbf{8.64} & \textbf{7.89} & \textbf{7.59} & \textbf{7.20} \\
      \bottomrule
      \end{tabular}%
    \caption{Comparison of METEOR score between our method based on Hard Binary Mask and Gaussian Mask on the samples with different overlap rate in the ActivityNet Caption validation set.}
    \label{tab:ratio}%
\end{table}% 

 \subsection{Impact of event overlap situation on our method.}
Sometimes, two events may have a overlap in time duration. We address this concern as follows.
First, from a technical perspective, our employed Gaussian mask allows partial semantic information of the masked area to be retained when an event is masked. The retained intensity gradually decreases from the center of the Gaussian mask toward both edges. As a result, even if two events overlap in time duration, the overlapping information can still be partially retained, and thus our negative masked captioning can still work.
Moreover, from an experimental perspective, the ActivityNet dataset contains many instances of event overlap, and our method outperforms existing methods on this dataset. We further divide the ActivityNet Caption validation set into four groups based on the average overlap rate between adjacent events in each video and list the results below. As shown in Table \ref{tab:ratio}, although the performance metrics decline as overlap increases, the decrease is not substantial compared to the hard binary mask.

\end{document}